\documentclass{article}

 \usepackage[preprint]{main}

\usepackage[utf8]{inputenc} 
\usepackage[T1]{fontenc}    
\usepackage{hyperref}      
\usepackage{url}            
\usepackage{booktabs}       
\usepackage{amsfonts}       
\usepackage{nicefrac}       
\usepackage{microtype}      
\usepackage{xcolor}      

\usepackage{multirow}
\usepackage{amsmath}
\usepackage{graphicx}
\usepackage{subcaption}
\usepackage{wrapfig}

\usepackage{enumitem}
\setlist[itemize]{leftmargin=*}

\title{FlowOVD: Learning Generative Latent Flows for Zero-shot Open-vocabulary Detection}

\author{
  Yao Wei$^{1}$, Andrea Cavallaro$^{2}$, Changjae Oh$^{1}$ \\
  \\
  $^{1}$Queen Mary University of London $^{2}$EPFL \\\\
  \texttt{https://qm-ipalab.github.io/FlowOVD/}
}

\newcommand{\modelname}{FlowOVD}

\begin{document}

\maketitle

\begin{abstract}

Open-vocabulary object detection (OVD) has achieved remarkable progress through large-scale vision-language pre-training. Existing methods, however, typically formulate OVD as a discriminative prediction problem, where decoder queries are either static or initialized from encoder features, thus limiting their diversity and flexibility. In this paper, we introduce a generative perspective by modeling decoder query generation as a continuous transport process in latent space. We propose \modelname{}, a text-conditioned query generation framework based on rectified flow that progressively transforms text-agnostic queries into text-guided queries. By introducing continuous latent query dynamics into a vision-language model (VLM) based detector, our method avoids heuristic discrete query construction and enables more expressive semantic alignment for open-vocabulary detection. Without requiring additional training data, \modelname{} achieves 49.5 AP on COCO and 31.5 AP on LVIS, outperforming GroundingDINO by +1.2 AP (+2.5 \%) and +4.1 AP (+15.0 \%), respectively. The larger gain on the challenging long-tailed LVIS benchmark further highlights the effectiveness of continuous query generation for open-vocabulary generalization.

\end{abstract}

\section{Introduction}
\label{intro}

\begin{figure}[t]
  \centering
  \includegraphics[width=\textwidth]{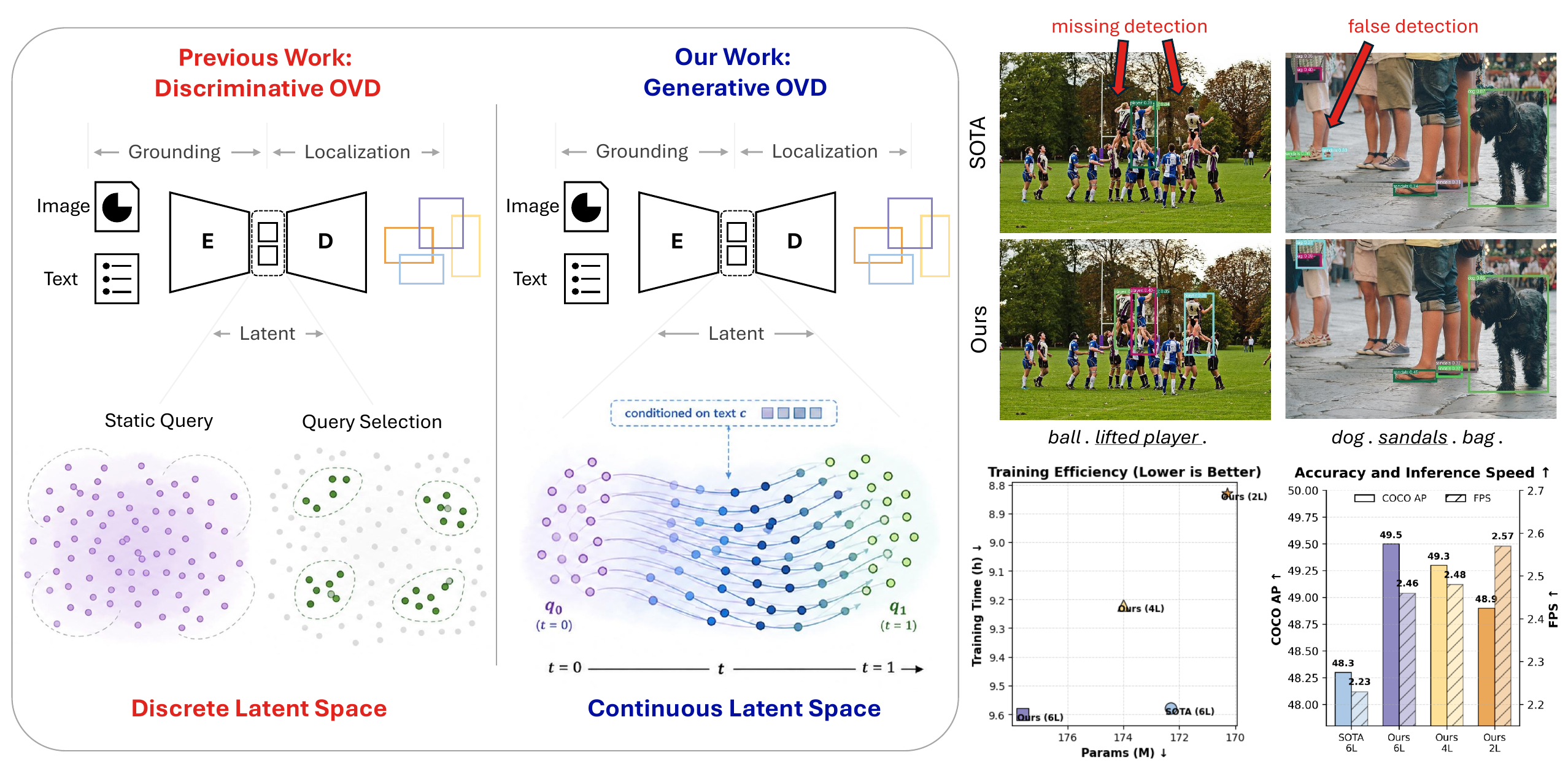}
  \caption{We model query initialization as a continuous, text-conditioned flow in latent space, transporting text-agnostic queries to text-guided queries, leading to more diverse and expressive open-vocabulary detection and better efficiency with fewer decoder layers.}
  \label{fig:teaser}
\end{figure}

Open-vocabulary object detection (OVD) aims to detect and localize objects beyond a predefined set of categories. Recent advances in vision-language models (VLMs) have substantially improved OVD performance by aligning visual and textual representations within Transformer-based architectures ~\cite{li2022grounded,liu2024grounding,lu2025dynamic}. In particular, methods such as GroundingDINO~\cite{liu2024grounding} demonstrate strong performance by integrating grounded pre-training with a strong Transformer detector and end-to-end optimization. Despite these advances, OVD approaches are fundamentally \emph{discriminative}, relying on explicit supervision signals such as contrastive learning for grounding and bounding box regression for localization. This reliance induces a discrete latent space, which hinders the modeling of fine-grained semantic variations. As shown in Fig.~\ref{fig:teaser}, object queries are typically initialized using static queries~\cite{zhang2022dino}, such as learnable embeddings, which remain the same for different images during inference. Alternatives are heuristics~\cite{liu2024grounding} like selecting Top$K$ proposals from encoder features. While effective, these designs impose a rigid structure on the query distribution, limiting its diversity and expressiveness. This restriction is problematic in open-vocabulary settings, where the model must handle a wide range of semantic concepts with varying visual appearances.

Generative modeling (e.g., ~\cite{ho2020denoising,liu2022flow} provides a principled way to capture underlying data distributions and naturally encourages diversity. Despite success in modeling complex visual distributions, the potential of generative modeling for discriminative vision-language tasks remains largely underexplored. Recent works~\cite{chen2023diffusiondet,baty2025flowdet} begin to incorporate generative processes into object detection by modeling distributions over object bounding boxes. While these approaches demonstrate the promise of generative modeling for localization tasks, they focus on geometric object generation without semantic representation learning. In contrast, OVD relies on cross-modal semantic alignment, where language conditions are supposed to actively guide the generation for open-world generalization.

We present \textbf{\modelname{}}, a novel OVD approach that reformulates query initialization as a continuous generative process in the latent space. We propose a \emph{text-conditioned query flow} that transforms an initial set of text-agnostic queries into a text-guided distribution. Specifically, we adopt a rectified flow formulation to model a time-dependent velocity field that transports queries under language conditioning. This enables controllable, and diverse query generation while remaining fully compatible with existing Transformer-based detectors. Our approach offers several key advantages. First, it enhances query diversity by exploring a continuous latent trajectory rather than relying on discrete query selection strategies. Second, it naturally incorporates text conditioning into the generative process, improving region-text alignment. Third, it introduces minimal architectural changes and can be seamlessly integrated into existing OVD pipelines. We evaluate our method on widely used benchmarks COCO and LVIS, where it consistently improves over strong baselines. These results demonstrate the effectiveness of incorporating generative modeling into query representations for OVD.
Our main contributions are summarized as follows:
\begin{itemize}
\item We reformulate decoder query generation as a continuous transport process in latent space, introducing a generative perspective for OVD and moving beyond conventional discrete query construction strategies.

\item We propose a text-conditioned query flow based on rectified flow, which progressively transforms text-agnostic queries into semantically aligned query representations and can be seamlessly integrated into existing Transformer-based vision-language detectors.

\end{itemize}

Extensive experiments on COCO and LVIS demonstrate that our approach consistently improves open-vocabulary detection performance and achieves better efficiency with fewer decoder layers.

\section{Related Works}
\label{literature}

\subsection{Open-vocabulary Object Detection}

Modern object detectors include CNN-based approaches such as the R-CNN series~\cite{ren2015faster,he2017mask} and YOLO series~\cite{redmon2016you,redmon2017yolo9000}, as well as Transformer-based approaches like DETR and its variants~\cite{carion2020end,zhu2020deformable,zhang2022dino}. Despite impressive performance, these detectors operate under a closed-set assumption, which restricts their generalization to novel categories in real-world scenarios. 

OVD addresses this limitation by leveraging external semantic knowledge to detect objects beyond predefined categories. One key challenge in OVD is aligning visual representations with language embeddings, enabling models to transfer knowledge from large-scale vision-language pre-training. Recent advances in detection architectures, such as DyHead~\cite{dai2021dynamic} and DINO~\cite{zhang2022dino}, provide strong foundations for open-vocabulary detectors. Early vision-language detection frameworks such as GLIP~\cite{li2022grounded} demonstrate the effectiveness of grounding-based pre-training for open-vocabulary transfer. Building upon DETR-style architectures, GroundingDINO~\cite{liu2024grounding} further unifies detection and grounding within a Transformer-based framework through cross-modal feature interaction and region-text alignment, enabling direct localization using arbitrary text prompts. More recently, YOLO-World~\cite{cheng2024yolo} explores efficient real-time OVD, while GroundingDINO 1.5~\cite{ren2024grounding} and Dynamic-DINO~\cite{lu2025dynamic} further improve performance through model scaling and MoE-based adaptation.

However, the modeling of latent query representations is largely simplified or overlooked in existing OVD frameworks. In Transformer-based detectors~\cite{carion2020end,zhu2020deformable,zhang2022dino,liu2024grounding}, decoder queries are typically constructed either from static learnable embeddings or directly selected from encoder features through heuristic selection strategies. While such query construction mechanisms are effective, they may restrict the diversity and flexibility, especially in open-vocabulary scenarios requiring rich semantic generalization. In contrast, we introduce a generative perspective by modeling a continuous transport process in latent space, enabling more expressive and text-aligned query representations.

\subsection{Generative Modeling}

Generative models stand out for their ability to model the underlying data distribution. Representative architectures include Generative Adversarial Networks (GANs)~\cite{goodfellow2014generative}, Normalizing Flows~\cite{rezende2015variational}, Denoising Diffusion Models~\cite{ho2020denoising}, and Flow Matching~\cite{lipman2022flow}. Among them, diffusion models have achieved remarkable success due to their stable training and high-quality generation. More recently, flow-based methods, such as Rectified Flow~\cite{liu2022flow}, have emerged as an efficient alternative by directly learning continuous transformations between distributions. 

Several recent works have explored generative formulations for discriminative tasks, such as object detection~\cite{chen2023diffusiondet,baty2025flowdet} and referring video segmentation~\cite{wang2025deforming}. These works demonstrate the potential of generative modeling for improving diversity in visual localization tasks. Nevertheless, directly extending these works to OVD remains challenging. In particular, FlowDet~\cite{baty2025flowdet}, although conceptually related to our work, fundamentally operates in the bounding box space by learning a transport process from randomly sampled boxes toward ground-truth box distributions. This formulations mainly focus on generative geometric localization, while the core challenge of OVD lies in visual-text alignment, where language conditions should effectively guide the generation process for semantic generalization. Moreover, box-space generative detectors typically require pre-processing like box padding due to varying numbers of objects across diverse images, which may reduce flexibility. 

Motivated by these observations, we introduce generative modeling into the latent query space of Transformer-based OVD frameworks. Instead of operating directly in box space as in~\cite{chen2023diffusiondet,baty2025flowdet}, our method models a continuous, text-conditioned flow over semantic query representations, enabling more controllable and semantically aligned query generation while naturally preserving compatibility with dominant vision-language detection frameworks.

\section{Method}
\label{method}

\begin{figure}[t]
  \centering
  \includegraphics[width=\textwidth]{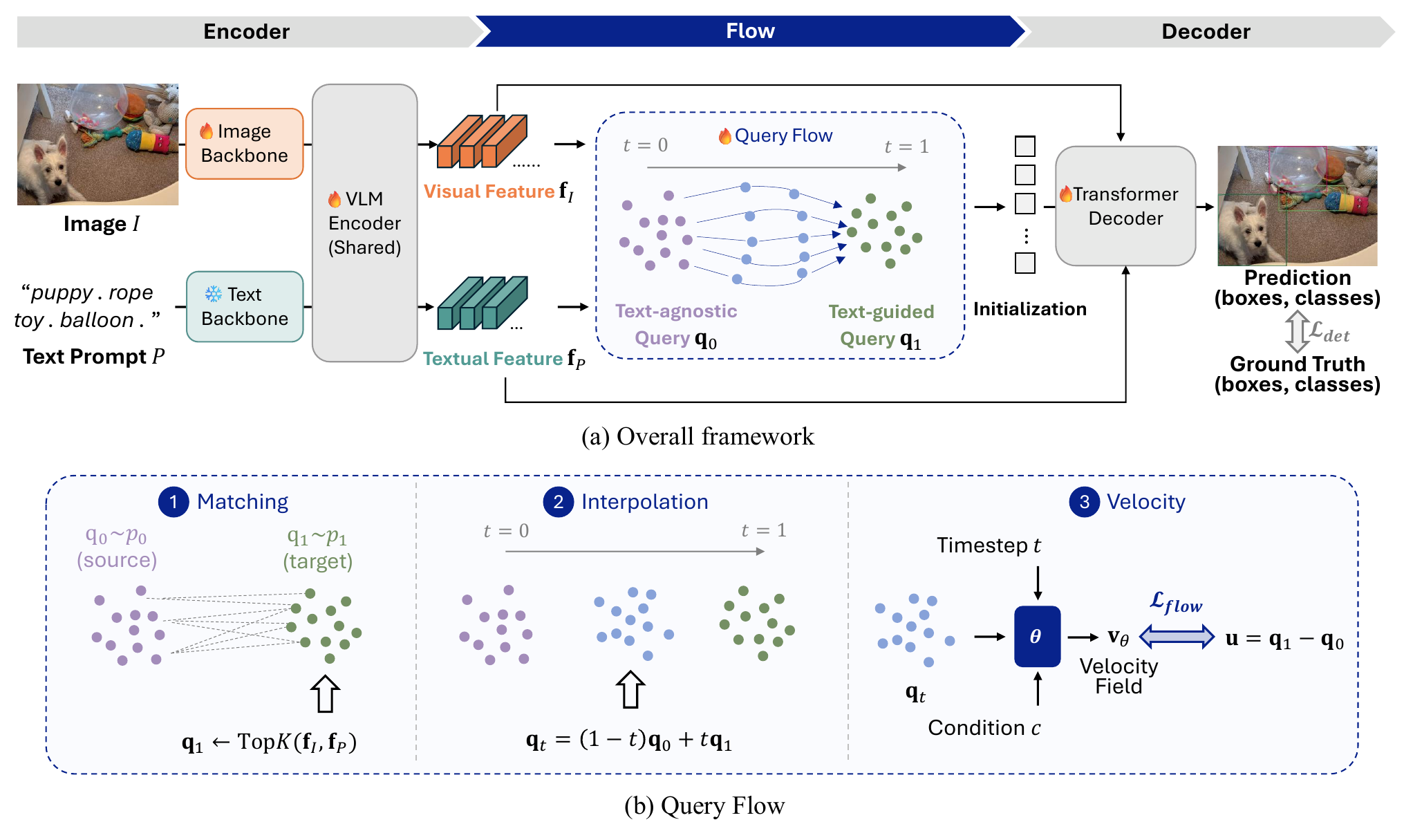}
  \caption{(a) \textbf{Overall framework.} Given an image $I$ and a text prompt $P$, a vision-language encoder extracts visual features $\mathbf{f}_I$ and textual features $\mathbf{f}_P$. In the latent space, we present a query flow that transforms a set of text-agnostic queries into text-conditioned queries. The refined queries are then consumed by the transformer decoder to produce final predictions. (b) \textbf{Query Flow.} First, a set-level matching is performed between source queries $\mathbf{q}_0$ and target queries $\mathbf{q}_1$. Intermediate states $\mathbf{q}_t$ are obtained via linear interpolation. A velocity field $\mathbf{v}_\theta(\mathbf{q}_t, t, c)$ is trained to learn the transport dynamics with conditions $c$. The final queries are obtained by integrating the learned flow, resulting in more diverse and text-aligned query representations.}
  \label{fig:pipeline}
\end{figure}

\subsection{Overview}

Traditionally, OVD is formulated as a discriminative prediction task through a Transformer encoder-decoder architecture~\cite{liu2024grounding}. Given an input image $I$ and a text prompt $P$, visual and textual backbones are first employed to extract visual features and textual embeddings, respectively. These features are subsequently fused via a shared vision-language encoder, enabling cross-modal interaction between image regions and textual tokens. A set of object queries is then used by the Transformer decoder to produce object-wise predictions, including bounding boxes $\mathbf{B}=\{\mathbf{b}_1,\mathbf{b}_2,\dots,\mathbf{b}_N\}$ and corresponding class labels $\mathbf{C}=\{\mathbf{c}_1,\mathbf{c}_2,\dots,\mathbf{c}_N\}$, where $N$ represents the number of queries. Despite its effectiveness, such a formulation relies on discrete (e.g., static) query initialization, which limits the diversity and flexibility of query representations. 

As illustrated in Fig.~\ref{fig:pipeline} (a), our \modelname{} builds upon a DETR-like architecture and consists of three components: a vision-language encoder, a query flow module in the latent space, and a Transformer decoder. We first construct an initial set of $N$ text-agnostic queries $\mathbf{q}_0 \in \mathbb{R}^{N \times d}$, which can be implemented as learnable embeddings, inspired by Rectified Flow~\cite{liu2022flow}. Meanwhile, we obtain a set of text-guided target queries $\mathbf{q}_1 \in \mathbb{R}^{N \times d}$ by performing query selection over encoder features $\mathbf{f}_I$ and $\mathbf{f}_P$. Instead of directly feeding either $\mathbf{q}_0$ or $\mathbf{q}_1$ into the decoder as previous works (e.g., ~\cite{zhang2022dino,liu2024grounding}), we introduce \textit{Query Flow} that learns a continuous transformation, which is achieved by a velocity network (parameterized as $\theta$), between text-agnostic and text-guided query distributions. The goal is to gradually refine queries toward semantically aligned representations. The resulting queries of the learned Query Flow, denoted as $\mathbf{q}_{\text{flow}}$, are then passed to the Transformer decoder to produce final detection results. By re-framing query initialization as a conditional generative process, our method enhances the diversity of query representations while preserving strong alignment with textual semantics, which naturally fits with OVD in a zero-shot setting.

\subsection{Text-conditioned Query Flow}

To enable more diverse latent query representations, we introduce text-conditioned Query Flow that learns a continuous transport process from an initial query set $\mathbf{q}_0 \sim p_0$ to a target query set $\mathbf{q}_1 \sim p_1$. Here, $p_0$ denotes a source distribution of text-agnostic queries (e.g., learnable embeddings), while $p_1$ is a target distribution of text-guided queries constructed from encoder features. Specifically, we follow a language-guided query selection strategy~\cite{liu2024grounding} to obtain $\mathbf{q}_1$. Given visual features $\mathbf{f}_I \in \mathbb{R}^{N_I \times d}$ and textual features $\mathbf{f}_P \in \mathbb{R}^{N_P \times d}$, we first compute a similarity matrix:
\begin{equation}
\mathbf{S} = \mathbf{f}_I \mathbf{f}_P^\top,
\end{equation}
where $\mathbf{S} \in \mathbb{R}^{N_I \times N_P}$ measures the alignment between image tokens and text tokens. Then, the similarity is aggregated over text tokens to obtain a score for each image token. The target query $\mathbf{q}_1$ is constructed by selecting the top $K$ image tokens with the highest scores:
\begin{equation}
\mathbf{q}_1 = \mathbf{f}_I\left[k\right], \quad
k=\text{Top}K\!\left( \max_{j} \mathbf{S}_{:,j} \right).
\end{equation}

As shown in Fig.~\ref{fig:pipeline} (b), Query Flow contains the following key concepts.

\textbf{Matching.} Since $\mathbf{q}_0$ and $\mathbf{q}_1$ are unordered sets, we first establish a one-to-one correspondence between them. 
We perform set-level matching (e.g., greedy matching based on cosine similarity) to obtain paired queries, ensuring that each source query in $\mathbf{q}_0$ is aligned with a corresponding target query in $\mathbf{q}_1$. 
This matching step provides supervision for learning a meaningful transport trajectory in the latent space.

\textbf{Interpolation.} Given matched pairs $(\mathbf{q}_0, \mathbf{q}_1)$, we construct intermediate states via linear interpolation:
\begin{equation}
\mathbf{q}_t = (1 - t)\mathbf{q}_0 + t \mathbf{q}_1,
\end{equation}

where $t \in [0,1]$ is a timestep. This interpolated $\mathbf{q}_t$ defines the transport path connecting the source and target query distributions.

\textbf{Velocity.} We model the transformation using a time-dependent velocity field $\mathbf{v}_\theta(\mathbf{q}_t, t, c) \in \mathbb{R}^{N \times d}$, where $c$ denotes the textual condition (i.e., $\mathbf{f}_P$) derived from encoder features. Formally, the query evolution follows an ordinary differential equation (ODE). The dynamics can be formulated as:
\begin{equation}
\frac{d\mathbf{q}_t}{dt} = \mathbf{v}_\theta(\mathbf{q}_t, t, c).
\end{equation}

\begin{figure}[t]
  \centering
  \includegraphics[width=.9\textwidth]{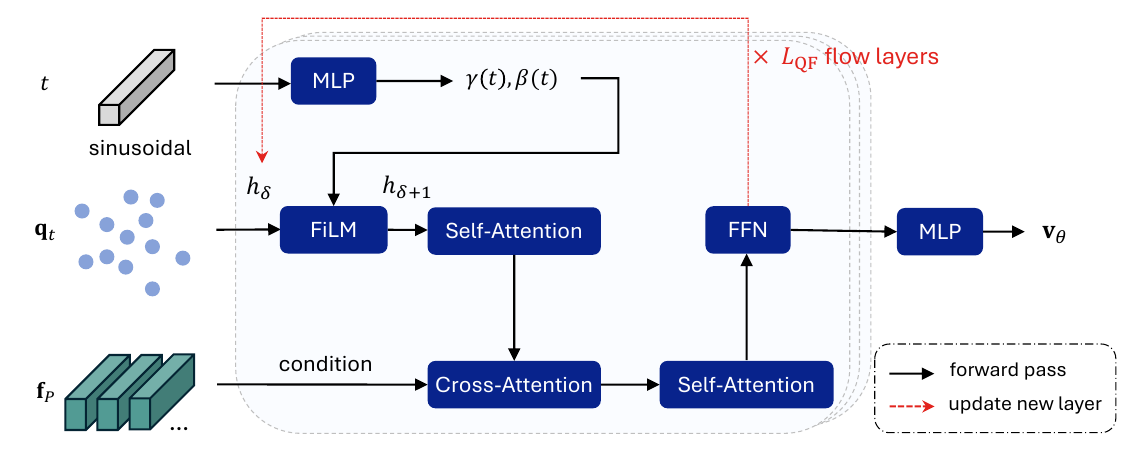}
  \caption{The architecture of velocity network $\theta$, which contains $L_{\text{QF}}$ flow layers. Taking timestep $t$, interpolated query $\mathbf{q}_t$, and textual feature $\mathbf{f}_P$ (condition) as input, $\theta$ learns to predict velocity field $\mathbf{v}_\theta$.}
  \label{fig:flow}
\end{figure}

As depicted in Fig.~\ref{fig:flow}, the velocity network is achieved by a lightweight Transformer. Textual conditioning is incorporated via cross-attention to text features. In addition, the timestep $t$ is encoded using sinusoidal positional encoding~\cite{vaswani2017attention}, followed by multi-layer perceptron (MLP) layers to yield modulation parameters, scale $\gamma(t)$ and shift $\beta(t)$. We adopt Feature-wise Linear Modulation (FiLM)~\cite{perez2018film} to inject time embeddings into intermediate representations $h$:
\begin{equation}
h_{\delta+1} = (\gamma(t)+1) \cdot h_{\delta} + \beta(t), 
\quad \delta = 0, 1, 2, \dots, L_{\text{QF}}-1.
\end{equation} 

where $h_{\delta}$ is initialized from the interpolated queries $\mathbf{q}_t$, and $L_{\text{QF}}$ is the number of flow layers. Each layer consists of a self-attention layer, a text cross-attention layer, another self-attention layer, and a feed-forward network (FFN), with residual connections and layer normalization applied after each sub-layer. Stacking $L_{\text{QF}}$ such layers forms the complete velocity network. This design enables the model to jointly capture intra-query interactions and query-text alignment. 

\subsection{Training Objective}

We adopt the flow matching objective, which supervises the velocity network $\theta$ to approximate the transport path between source and target distributions. The ground-truth velocity field is defined as:
\begin{equation}
\mathbf{u} = \mathbf{q}_1 - \mathbf{q}_0.
\end{equation}

The flow loss is computed by the mean squared error between the predicted velocity and the ground-truth transport direction:
\begin{equation}
\mathcal{L}_{\text{flow}} = 
\mathbb{E}_{t, \mathbf{q}_0, \mathbf{q}_1} 
\left[
\left\| \mathbf{v}_\theta(\mathbf{q}_t, t, c) - \mathbf{u} \right\|_2^2
\right],
\end{equation}

\begin{figure}[t]
    \centering
    \begin{subfigure}{0.24\textwidth}
        \centering
        \includegraphics[width=\linewidth]{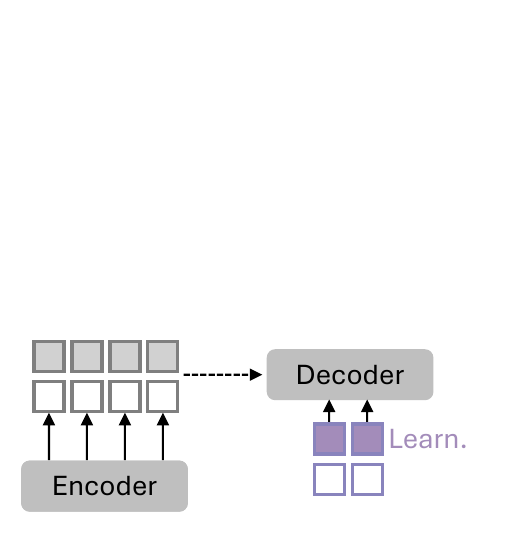}
        \caption{Static Query~\cite{carion2020end}}
        \label{fig:4-1}
    \end{subfigure}
    \hfill
    \begin{subfigure}{0.24\textwidth}
        \centering
        \includegraphics[width=\linewidth]{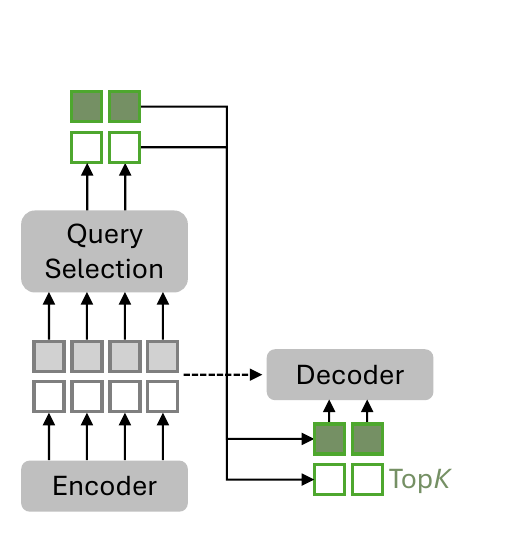}
        \caption{Vanilla Selection~\cite{zhu2020deformable}}
        \label{fig:4-2}
    \end{subfigure}
    \hfill
    \begin{subfigure}{0.24\textwidth}
        \centering
        \includegraphics[width=\linewidth]{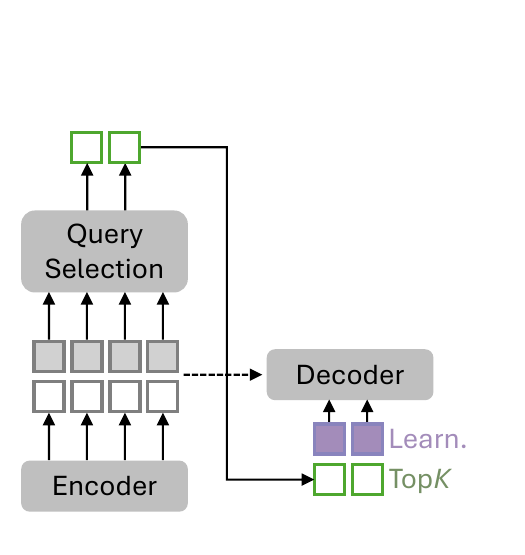}
        \caption{Mixed Selection~\cite{zhang2022dino}}
        \label{fig:4-3}
    \end{subfigure}
    \hfill
    \begin{subfigure}{0.24\textwidth}
        \centering
        \includegraphics[width=\linewidth]{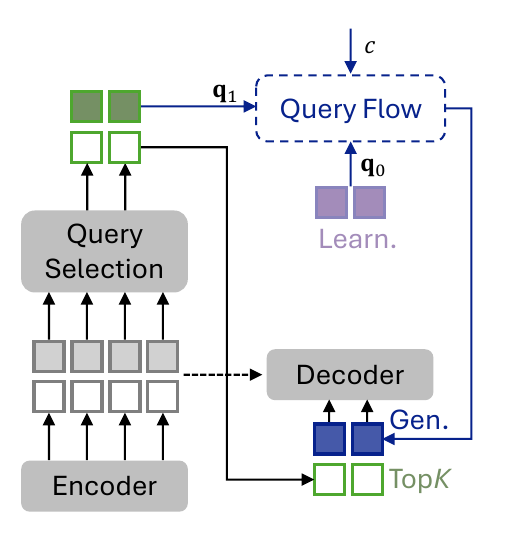}
        \caption{Query Flow (Ours)}
        \label{fig:4-4}
    \end{subfigure}

    \caption{Different query construction strategies for initializing decoder queries, where filled and hollow squares are content queries and positional queries, respectively. KEY-- Learn.: learnable query embeddings, Gen.: generative content queries. Best viewed in color.}
    \label{fig:query_select}

\end{figure}

which is jointly optimized with the standard detection loss, including the bounding box regression loss~\cite{zhang2022dino} composed of L1 and GIOU losses~\cite{rezatofighi2019generalized}, and the classification loss~\cite{li2022grounded} computed by the focal loss~\cite{lin2017focal}:
\begin{equation}
\begin{aligned}
 \mathcal{L}_{\text{det}} &= \mathcal{L}_{\text{bbox}} + \mathcal{L}_{\text{class}} \\ &= \mathcal{L}_{\text{1}} + \mathcal{L}_{\text{GIOU}} + \mathcal{L}_{\text{focal}}.
\end{aligned}
\end{equation}

The total training objectives is:
\begin{equation}
\mathcal{L} = \mathcal{L}_{\text{det}} + \lambda \mathcal{L}_{\text{flow}},
\end{equation}

where $\lambda$ is the balancing coefficient of the flow objective. 

Overall, the detection loss enforces the discriminative localization capability, while the flow loss leads to generating semantically aligned query representations. Unlike box-space generative methods~\cite{chen2023diffusiondet,baty2025flowdet}, our model is trained in the latent space, avoiding explicit box generation and enabling seamless integration with Transformer-based VLMs. 

\subsection{Inference}

At inference time, we generate refined queries by integrating the learned text-conditioned velocity field starting from the initial query set $\mathbf{q}_0$. The transported queries are obtained by integrating the velocity field from $t=0$ to $t=1$:
\begin{equation}
\mathbf{q}_{\text{flow}} = \mathbf{q}_0 + \int_0^1 \mathbf{v}_\theta(\mathbf{q}_t, t, c) \, dt.
\end{equation}

In practice, this continuous process is approximated using Euler discretization with step size $\epsilon$. Starting from $\mathbf{q}_0$, the queries are iteratively updated as:
\begin{equation}
\mathbf{q}_{t+\epsilon}=\mathbf{q}_t+\epsilon \cdot \mathbf{v}_\theta(\mathbf{q}_t, t, \mathbf{c}),
\end{equation}

where $t \in [0,1]$ is uniformly increased by $\epsilon$ at each integration step. After several iterations, the final transported state is taken as $\mathbf{q}_{\text{flow}}$.

To balance stability and expressiveness, we combine the transported queries with the original queries:
\begin{equation}
\mathbf{q}_{\text{init}} = (1 - \alpha)\mathbf{q}_0 + \alpha \mathbf{q}_{\text{flow}},
\end{equation}
where $\alpha$ controls the strength of the generative refinement.

In Transformer-based detectors, each query consists of two components: a \emph{content query} and a \emph{positional query}. The content query encodes semantic information for feature interaction, while the positional query provides spatial priors (e.g., reference points) for object localization. In this work, we apply the proposed flow-based refinement only to the content queries, while keeping the positional queries unchanged. The refined content queries $\mathbf{q}_{\text{init}}$, together with the original positional queries, are then fed into the Transformer decoder for final object prediction.

Fig.~\ref{fig:query_select} illustrates the evolution of query initialization strategies. Existing approaches either rely on static learnable queries (Learn.), pure Top$K$ selection from encoder features, or a heuristic combination of both. In contrast, our method introduces a flow-based query refinement module that transports learnable queries $\mathbf{q}_0$ toward text-guided target queries $\mathbf{q}_1$ in a continuous manner. This reformulates query initialization as a generative process in the latent space, leading to more flexible, diverse, and text-aligned query representations (Gen.).

\begin{table}[t]
  \caption{Zero-shot domain transfer on COCO. AP$_s$, AP$_m$, and AP$_l$ are the AP of small, medium, and large categories, respectively. \dag indicates the results of our replication, which is also used for our model's initialization. Datasets shown in \textcolor{gray}{gray} are only used for model initialization. }
  \label{tab:coco}
  \centering
  \resizebox{\linewidth}{!}{
  \begin{tabular}{llll|ll}
    \toprule
    \multirow{2}{*}{\textbf{Method}}     & \multirow{2}{*}{\textbf{Backbone}}  & \multirow{2}{*}{\textbf{Pre-Training Data}}  & \multirow{2}{*}{\textbf{Size}} & \multicolumn{2}{c}{\textbf{COCO} \texttt{val2017}} \\ 
    & & & & AP & AP$_s$ / AP$_m$ / AP$_l$\\
    \midrule
    DyHead~\cite{dai2021dynamic} & Swin-T & O365 & 0.61M & 43.6 & - \\
    GLIP~\cite{li2022grounded} & Swin-T & O365 & 0.61M & 44.9 & - \\
    DINO~\cite{zhang2022dino} & Swin-T & O365 & 0.61M & 46.2 & - \\
    GroundingDINO~\cite{liu2024grounding} & Swin-T & O365 & 0.61M & 46.7  & - \\
    GroundingDINO \dag & Swin-T & O365 & 0.61M  & 45.3  & 31.6 / 48.1 / 57.6 \\
    \modelname{} (Ours) & Swin-T & O365 & 0.61M & 45.5 & 32.0 / 48.4 / 58.0 \\
    \midrule
    YOLO-World \cite{cheng2024yolo} & YOLOv8 & O365,GoldG,CC3M & 1.63M & 45.1 & - \\
    Dynamic-DINO~\cite{lu2025dynamic} & EfficientViT & O365,GoldG,V3Det & 1.58M & 46.2 & - \\
    GLIP~\cite{li2022grounded} & Swin-T & O365,GoldG,Cap4M & 5.38M & 46.3 & - \\
    GroundingDINO~\cite{liu2024grounding} & Swin-T & O365,GoldG,Cap4M & 5.38M & 48.3  & 34.0 / 51.6 / 62.9 \\
    \modelname{} (Ours) & Swin-T & O365\textcolor{gray}{,GoldG,Cap4M}  & 0.61M & \textbf{49.5} & \textbf{35.4} / \textbf{52.6} / \textbf{63.6} \\
    \bottomrule
  \end{tabular}}
\end{table}

\begin{table}[t]
  \caption{Zero-shot domain transfer on LVIS. AP$_r$, AP$_c$, and AP$_f$ are the AP of rare, common, and frequent categories, respectively. \dag indicates the results of our replication, which is also used for our model's initialization.}
  \label{tab:lvis}
  \centering
  \begin{tabular}{lll|ll}
    \toprule
    \multirow{2}{*}{\textbf{Method}}  & \multirow{2}{*}{\textbf{Pre-Training Data}}  & \multirow{2}{*}{\textbf{Size}} & \multicolumn{2}{c}{\textbf{LVIS} \texttt{minival}} \\
    & & & AP & AP$_r$ / AP$_c$ / AP$_f$ \\
    \midrule
    GLIP~\cite{li2022grounded} & O365,GoldG & 1.38M & 24.9 & 17.7 / 19.5 / 31.0 \\
    GroundingDINO~\cite{liu2024grounding} & O365,GoldG  & 1.38M &  25.6  & 14.4 / 19.6 / 32.2 \\
    GLIP~\cite{li2022grounded} & O365,GoldG,Cap4M  & 5.38M & 26.0 & 20.8 / 21.4 / 31.0 \\
    GroundingDINO~\cite{liu2024grounding} & O365,GoldG,Cap4M  & 5.38M & 27.4  & 18.1 / 23.3 / 32.7 \\
    GroundingDINO \dag & O365,GoldG,Cap4M  & 5.38M & 24.3 & 17.2 / 20.9 / 28.6 \\
    \modelname{} (Ours) & O365\textcolor{gray}{,GoldG,Cap4M}  & 0.61M & \textbf{31.5} & \textbf{22.4} / \textbf{27.0} / \textbf{37.1} \\
    \bottomrule
  \end{tabular}
\end{table}

\section{Experiments}
\label{exp}

\subsection{Experimental Setup}

\noindent\textbf{Pre-training and evaluation datasets.} Our \modelname{} is trained on Objects365 (O365)~\cite{shao2019objects365}, a large-scale detection dataset containing over 600K training images. 
To assess zero-shot OVD performance, the pre-trained model is evaluated on COCO \texttt{val2017}~\cite{lin2014microsoft}, COCO with refined masks (COCO-ReM)~\cite{singh2024benchmarking}, and LVIS~\cite{gupta2019lvis}.  Following standard protocols, we report Average Precision (AP).

\noindent\textbf{Implementation details.} For the backbones, we consider the Swin Transformer Tiny (Swin-T) \cite{liu2021swin} as the visual backbone, and the BERT \cite{devlin2019bert} as the textual backbone. We set the number of flow layers $L_{\text{QF}}$ to 2 and the step size $\epsilon$ to 0.25. During training, the weight allocated to flow loss, i.e., $\lambda$, is set as 0.05. In addition, we set $\alpha=$ 0.2 to balance between learnable and generative queries when initializing decoder queries. The number of object queries $N$ is set to 900, with 6 decoder layers by default. To obtain $\mathbf{q}_1$, we set $K=N=$ 900. Our \modelname{} is trained on 4 NVIDIA H100 GPUs with a batch size of 16. We adopt the AdamW optimizer~\cite{loshchilov2017decoupled} with a base learning rate of $1 \times 10^{-4}$ and a specific learning rate of $2 \times 10^{-5}$ for backbones. More details are provided in Appendix.

\subsection{Results and Analysis}

Our \modelname{} is compared with the state-of-the-art open-vocabulary detectors, including DyHead~\cite{dai2021dynamic}, GLIP~\cite{li2022grounded}, DINO~\cite{zhang2022dino}, YOLO-World \cite{cheng2024yolo}, GroundingDINO~\cite{liu2024grounding}, and Dynamic-DINO~\cite{lu2025dynamic}.

As reported in Table~\ref{tab:coco}, the models are evaluated under two pre-training settings. For the standard O365-only setting, our method achieves 45.5 AP on COCO \texttt{val2017}, outperforming the reproduced GroundingDINO baseline by +0.2 AP while consistently improving performance across different object scales. When initialized from large-scale vision-language pre-training on O365, GoldG~\cite{kamath2021mdetr}, and Cap4M~\cite{li2022grounded}, \modelname{} further improves to 49.5 AP, surpassing GroundingDINO by +1.2 AP. The larger gain under stronger pre-training suggests that the proposed text-conditioned query flow can effectively leverage rich cross-modal semantic priors for query generation.

Table ~\ref{tab:lvis} presents the zero-shot transfer results on LVIS containing over 1200 categories, which is considerably more challenging due to its long-tailed category distribution. The reproduced baseline achieves 24.3 AP, where the difference from the reported results mainly comes from variations in prompt construction and evaluation settings. Under the same protocol, the proposed \modelname{} consistently improves over the baseline across all category frequencies. When initialized from large-scale vision-language pre-training on O365, GoldG, and Cap4M, \modelname{} achieves 31.5 AP on LVIS \texttt{minival}, outperforming GroundingDINO by +4.1 AP. Notably, the improvement is especially significant on rare categories, where AP$_r$ improves from 17.2 to 22.4. 

\begin{table}[t]
  \caption{Ablation study where all models are trained on the O365 dataset with a Swin-T backbone. The results in brackets are evaluated on COCO-ReM.}
  \label{tab:ablation}
  \centering
  \begin{tabular}{c|cccc}
    \toprule
    \multirow{2}{*}{\textbf{Query Initialization}}  & \multicolumn{4}{c}{\textbf{COCO (COCO-ReM)}} \\
    & AP & AP$_s$ & AP$_m$  & AP$_l$  \\
    \midrule
    Vanilla Selection & 44.9 (46.4) & 31.3 (33.1) & 47.9 (50.7) & 56.5 (59.9) \\
    Mixed Selection & 45.3 (46.7) & 31.6 (33.1) & 48.1 (50.9) & 57.6 (60.9)  \\
    Query Flow & \textbf{49.5 (51.1)} & \textbf{35.4 (36.6)} & \textbf{52.6 (55.7)} & \textbf{63.6 (67.4)} \\
    \bottomrule
  \end{tabular}
\end{table}

\begin{table}[t]
\centering
\caption{Efficiency comparisons. All models are trained on the O365 dataset with a Swin-T backbone using 4 H100 GPUs. We report model size (parameters), training time per epoch, and inference speed (FPS). Inference speed is measured on a single H100 GPU.}
\label{tab:decoder_efficiency}
\begin{tabular}{l|cc|cc|cc}
\toprule
\multirow{2}{*}{\textbf{Method}}
& \multicolumn{2}{c|}{\textbf{Configuration}} 
& \multicolumn{2}{c|}{\textbf{Training}} 
& \multicolumn{2}{c}{\textbf{Inference}} \\
& Flow & Decoder 
& Params (M) $\downarrow$ & Time (h) $\downarrow$ 
& COCO AP $\uparrow$ & FPS $\uparrow$ \\
\midrule
GroundingDINO & $\times$ & 6 layers & 172.3 & 9.58 & 48.3 & 2.23 \\
\midrule
\multirow{3}{*}{\modelname{}} & \checkmark & 6 layers  & 177.6 & 9.60 & \textbf{49.5} & 2.46 \\
& \checkmark & 4 layers & 174.0 & 9.22 & 49.3 & 2.48 \\
& \checkmark & 2 layers & \textbf{170.3} & \textbf{8.83} & 48.9 & \textbf{2.57} \\
\bottomrule
\end{tabular}
\end{table}

As demonstrated in Table~\ref{tab:ablation}, the proposed flow-based query generation consistently outperforms both vanilla and mixed query selection strategies across all evaluation metrics. These results suggest that reformulating query initialization as a continuous generative process in latent space leads to more expressive and semantically aligned query representations, which are especially beneficial for open-vocabulary generalization.

We analyze the efficiency of our method with respect to decoder depth in Table~\ref{tab:decoder_efficiency}. Compared with GroundingDINO, the proposed \modelname{} achieves both better efficiency and stronger detection performance. Specifically, our full 6-layer model improves COCO AP from 48.3 to 49.5 while introducing only a marginal increase in model size (177.6M vs. 172.3M). Despite the additional query flow formulation, the training time remains comparable, and inference speed is also slightly improved. Moreover, reducing the decoder depth improves both training and inference efficiency while maintaining competitive performance. For example, the lightweight 2-layer variant still outperforms GroundingDINO using substantially fewer parameters and higher inference speed.

\begin{figure}[t]
  \centering
  \includegraphics[width=.9\textwidth]{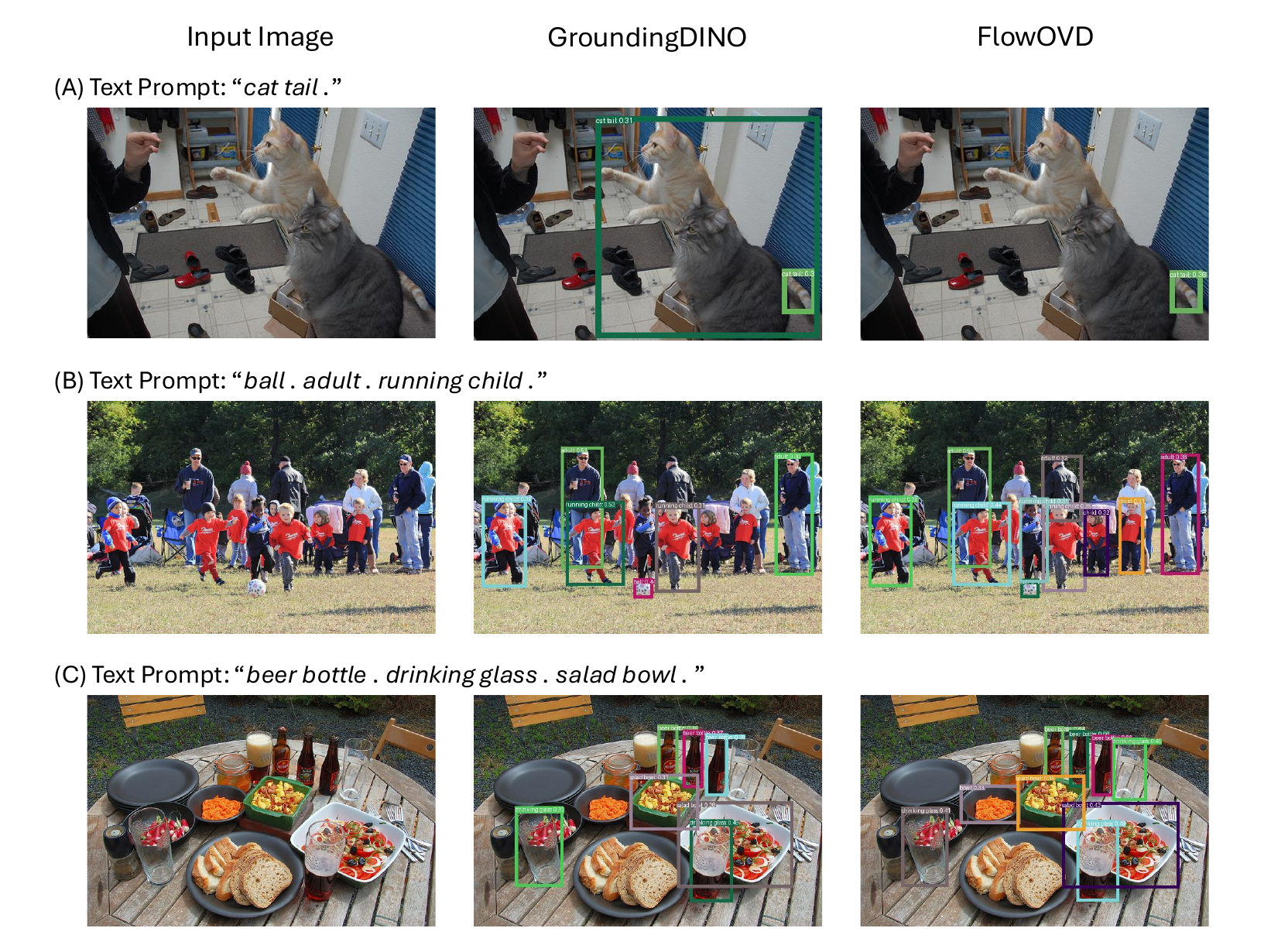}
  \caption{Qualitative comparison between GroundingDINO and our \modelname{} on COCO images. Given the same text prompts, \modelname{} demonstrates stronger grounding ability in fine-grained localization and compositional understanding.}
  \label{fig:result}
\end{figure}

Qualitative comparisons are provided in Figure~\ref{fig:result} to highlight the advantages of our method. As shown in Figure~\ref{fig:result} (A), GroundingDINO often responds to the entire object when queried with fine-grained phrases such as “cat tail”, indicating limited localization precision. In contrast, \modelname{} accurately focuses on the specified part, demonstrating improved alignment between textual semantics and visual regions. In Figure~\ref{fig:result} (B), GroundingDINO fails to detect certain queried concepts, including an occluded “running child” and an “adult”, while \modelname{} successfully identifies both, showing stronger robustness under occlusion. In particular, \modelname{} can further distinguish “running child” from a standing “child” on the right, indicating improved sensitivity to action-related language cues. In cluttered multi-object scenes (Figure~\ref{fig:result} (C)), \modelname{} achieves more complete detections across different object types compared to GroundingDINO. These results demonstrate that \modelname{} provides stronger grounding performance in open-vocabulary scenarios.

\section{Conclusion}
\label{conc}

We presented \modelname{}, a generative framework for OVD that reformulates decoder query initialization as a continuous, text-conditioned transport process in latent space. By learning a rectified flow from static text-agnostic queries to semantically aligned query representations, our method produces more diverse and expressive object queries for OVD. Extensive experiments on COCO and LVIS demonstrate consistent improvements over prior arts. Additionally, our flow-based query generation reduces the reliance on deep Transformer decoders, achieving improved efficiency while maintaining competitive detection performance. The results of our formulation demonstrate that generative latent modeling provides a promising alternative to discriminative query-based models for open-world vision-language perception.

\clearpage

\bibliographystyle{plain}
\bibliography{reference}

\appendix

\section{Additional Details}

\subsection{Dataset Details}

Our \modelname{} is evaluated on diverse benchmarks in a zero-shot setting. The COCO~\cite{lin2014microsoft} dataset is a widely used benchmark. It contains approximately 118,000 training images and 5,000 validation images with annotations from 80 object categories. In this work, we evaluate zero-shot transfer performance on the COCO \texttt{val2017} split using category names as text prompts. Moreover, COCO-ReM~\cite{singh2024benchmarking} is derived from COCO by removing overlapping categories between training and evaluation sets. Compared with standard COCO evaluation, COCO-ReM better measures the generalization ability of models to novel categories. The LVIS~\cite{gupta2019lvis} dataset, a more challenging benchmark, annotates COCO images~\cite{lin2014microsoft} with 2,000,000 instances among a set of 1,203 object categories. Following prior works~\cite{liu2024grounding}, we also employ the LVIS mini-validation (\texttt{minival}) for evaluation.

\subsection{Reproduction and Training Details}
We follow the default architecture and training settings of GroundingDINO~\cite{liu2024grounding}. Since the official training code is not publicly available, we reproduce the baseline based on the open-source implementation Open-GroundingDINO\footnote{https://github.com/longzw1997/Open-GroundingDino} with several modifications to better align with the original GroundingDINO training protocol. Specifically, we correct the computation of positive maps and revise the training curriculum by removing the early learning rate decay strategy, which we found limits the final detection performance.

During training, we freeze the textual backbone (i.e., BERT-base \cite{devlin2019bert}) while optimizing the visual backbone (i.e., Swin-T \cite{liu2021swin}) as well as the Query Flow and the detection decoder parameters end-to-end. For the standard O365-only setting, we train the model for 24 epochs and learning rate drops at the 15-th and 20-th epochs. For large-scale vision-language pre-training, the proposed Query Flow is trained from scratch, while the remaining detector weights are initialized from the checkpoint pretrained on O365, GoldG, and Cap4M. Specifically, the full model is further optimized end-to-end on O365 for 3 epochs. Table~\ref{tab:setting} details the major setting of hyperparameters.

\begin{table*}[h]
    \centering
    \small
    \caption{Hyperparameters for \modelname{}.
    }\label{tab:setting}
    \setlength{\tabcolsep}{10pt}
        \begin{tabular}{ll}
        \toprule
         \textbf{Hyperparameter}   &  \textbf{Value}  \\       
        \midrule
        \multicolumn{2}{c}{\footnotesize{\textit{General Configuration}}} \\
        \midrule
         GPUs    &   NVIDIA H100 $\times$ 4 \\
         batch size    &   16 \\
         optimizer    &   AdamW \\
         weight decay & $10^{-4}$ \\
         learning rate (base)  &  $1 \times 10^{-4}$  \\
         learning rate (backbone)  &  $2 \times 10^{-5}$  \\
         learning rate schedule & multi-step decay \\
         EMA decay    &   0.9997 \\
         image size & [800, 1333] \\
         encoder / decoder layers & 6 / 6 \\
         attention head & 8 \\
         matcher & Hungarian \\
         class / bbox / GIoU cost & 1.0 / 5.0 / 2.0 \\
         class / bbox / GIoU loss coef & 2.0 / 5.0 / 2.0 \\
         $K$ & 900 \\
         $N$ & 900 \\
         $d$ & 256 \\
         \midrule
         \multicolumn{2}{c}{\footnotesize{\textit{Flow Configuration}}} \\
        \midrule
         number of flow layers $L_{\text{QF}}$ & 2 \\
         step size $\epsilon$ & 0.25 \\
         strength of generative refinement $\alpha$ & 0.2 \\
         flow loss weight $\lambda$ & 0.05 \\
        \bottomrule
        \end{tabular}
    \end{table*}

\newpage
\subsection{Additional Qualitative Results}

\begin{figure}[h]
  \centering
  \includegraphics[width=\textwidth]{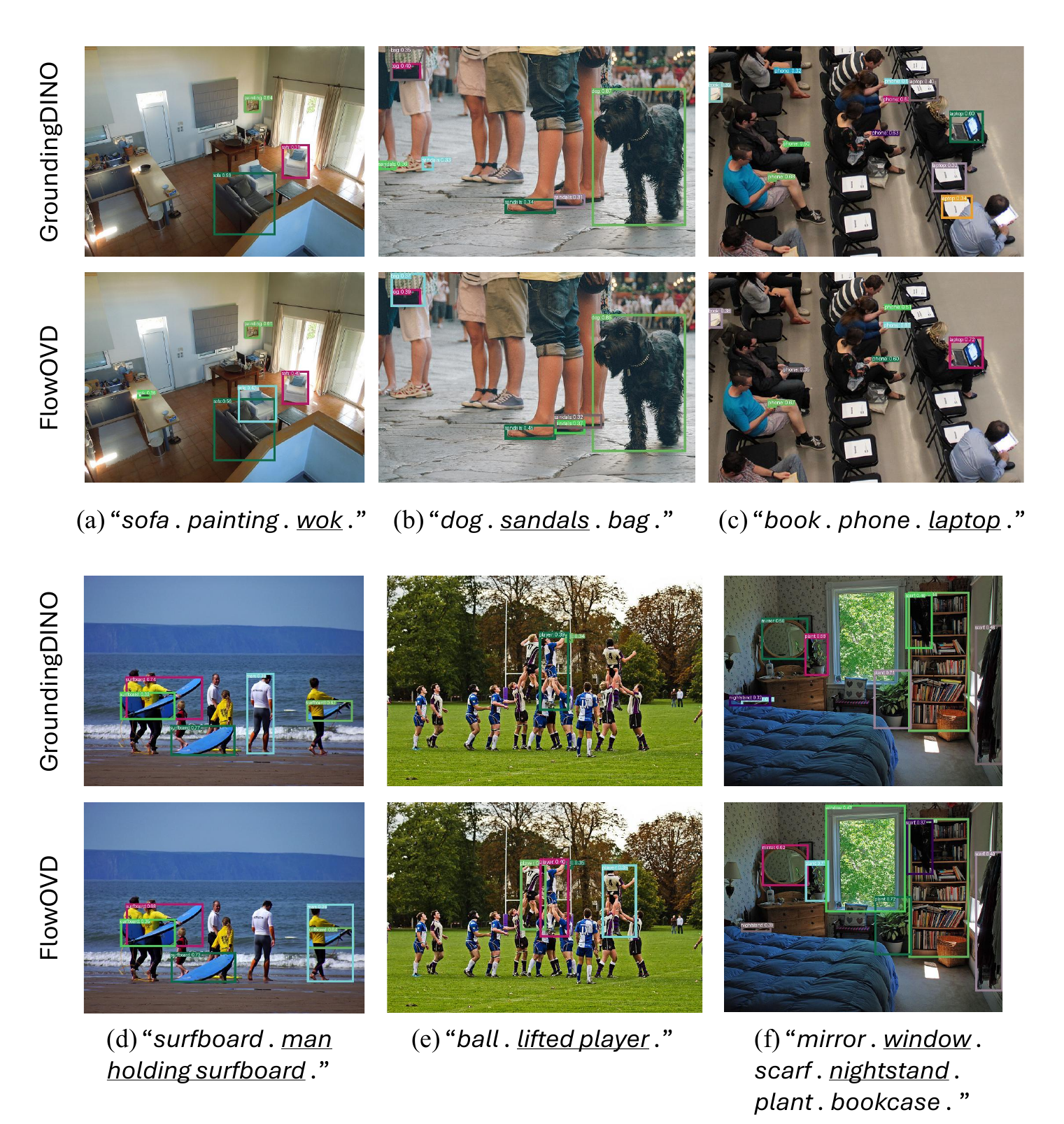}
  \caption{Additional qualitative comparisons. (a)(b): \modelname{} exhibits better generalization ability on rare categories such as \textit{wok} and \textit{sandals}. (c)(d): GroundingDINO often predicts false or repeated detection, for example, it fails to distinguish \textit{laptop} from booklets on chairs. In addition, \textit{nightstand} is repeatedly detected due to occlusion. (e)(f): When multiple instances of the same category coexist, \modelname{} demonstrates stronger semantic attribute-level discrimination, producing more accurate localization for compositional descriptions such as \textit{man holding surfboard} and \textit{lifted player}.}
  \label{fig:more_compare}
\end{figure}

Figure~\ref{fig:more_compare} provides additional qualitative comparisons which further demonstrate the advantages of our \modelname{} in open-vocabulary generalization. 

\begin{figure}[ht]
  \centering
  \includegraphics[width=\textwidth]{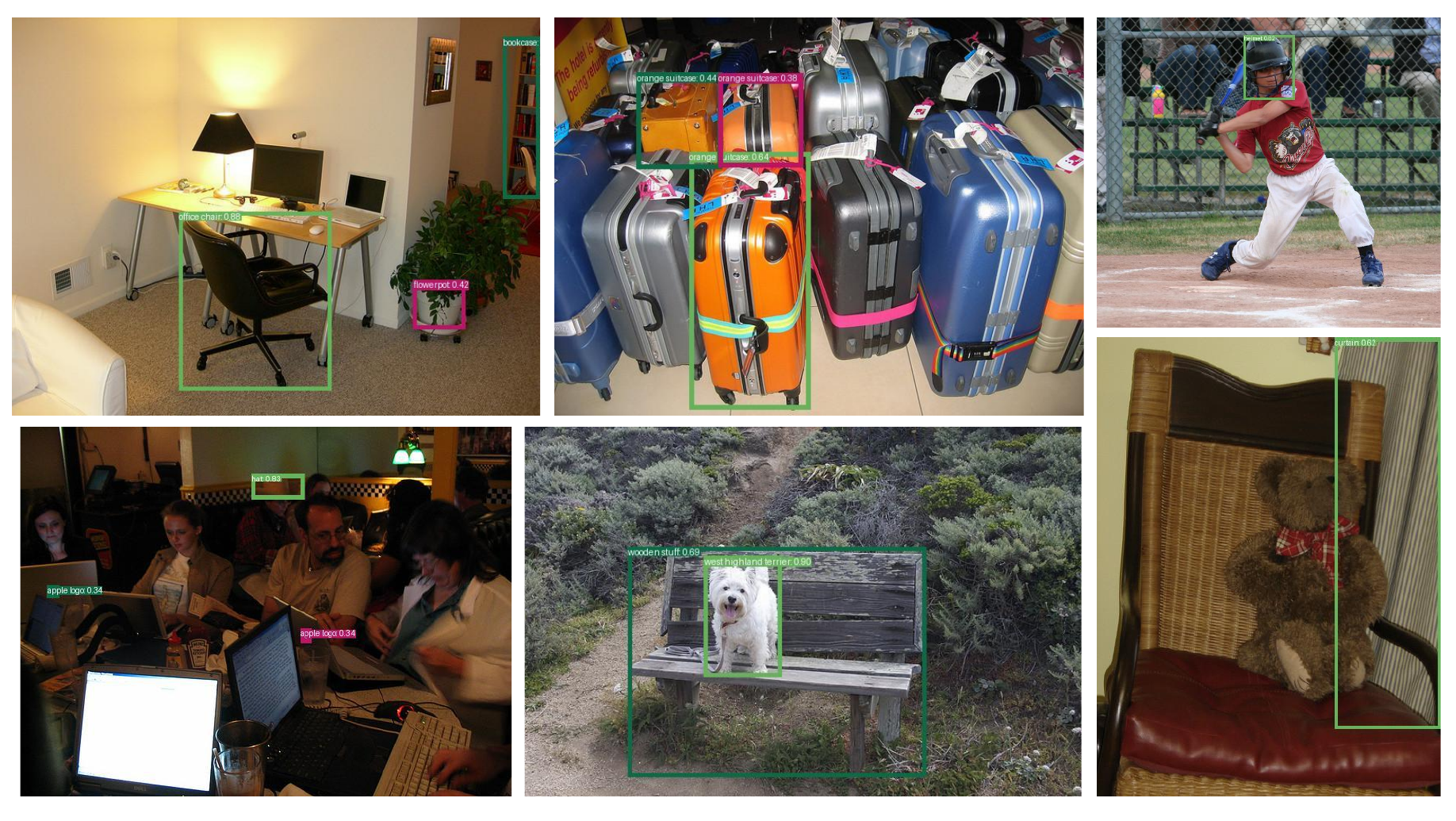}
  \caption{Additional examples of \modelname{}.}
  \label{fig:case}
\end{figure}

We also provide more zero-shot detection examples of our \modelname{} in Figure~\ref{fig:case}.

\newpage

\section{Discussions}

\subsection{Positional Query}

Our work focuses on learning a latent flow to the content queries, which mainly encodes semantic information for cross-modal interaction. In contrast, positional queries serve as spatial anchors that provide localization priors for the Transformer decoder. A natural question arises: why not directly apply flow-based generation to the entire query, including both content and positional queries?

To investigate this, we experimentally extend the proposed flow formulation to positional queries by modeling continuous transport in the positional latent space. However, we consistently observe degraded performance, for example, resulting in over 2 AP drops after only one epoch of training. We hypothesize that positional queries benefit from maintaining deterministic and stable spatial priors during decoder initialization. Continuous deformation in positional space may weaken the anchor behavior of positional queries, making decoder optimization less stable.

Furthermore, this observation highlights a key challenge in directly introducing box-space generative formulations (e.g.,~\cite{baty2025flowdet}) into OVD. Although geometric generation in box or positional space appears more intuitive for localization, OVD fundamentally relies on semantic alignment between visual regions and language descriptions. In contrast, content queries encode high-level semantic representations and naturally exhibit greater diversity in open-vocabulary settings, making them more suitable for continuous generative modeling. Applying flow-based refinement to content queries thus enhances semantic flexibility and expressiveness while preserving the stable localization priors provided by positional queries.

\subsection{Failure Cases}

\begin{figure}[ht]
  \centering
  \includegraphics[width=.9\textwidth]{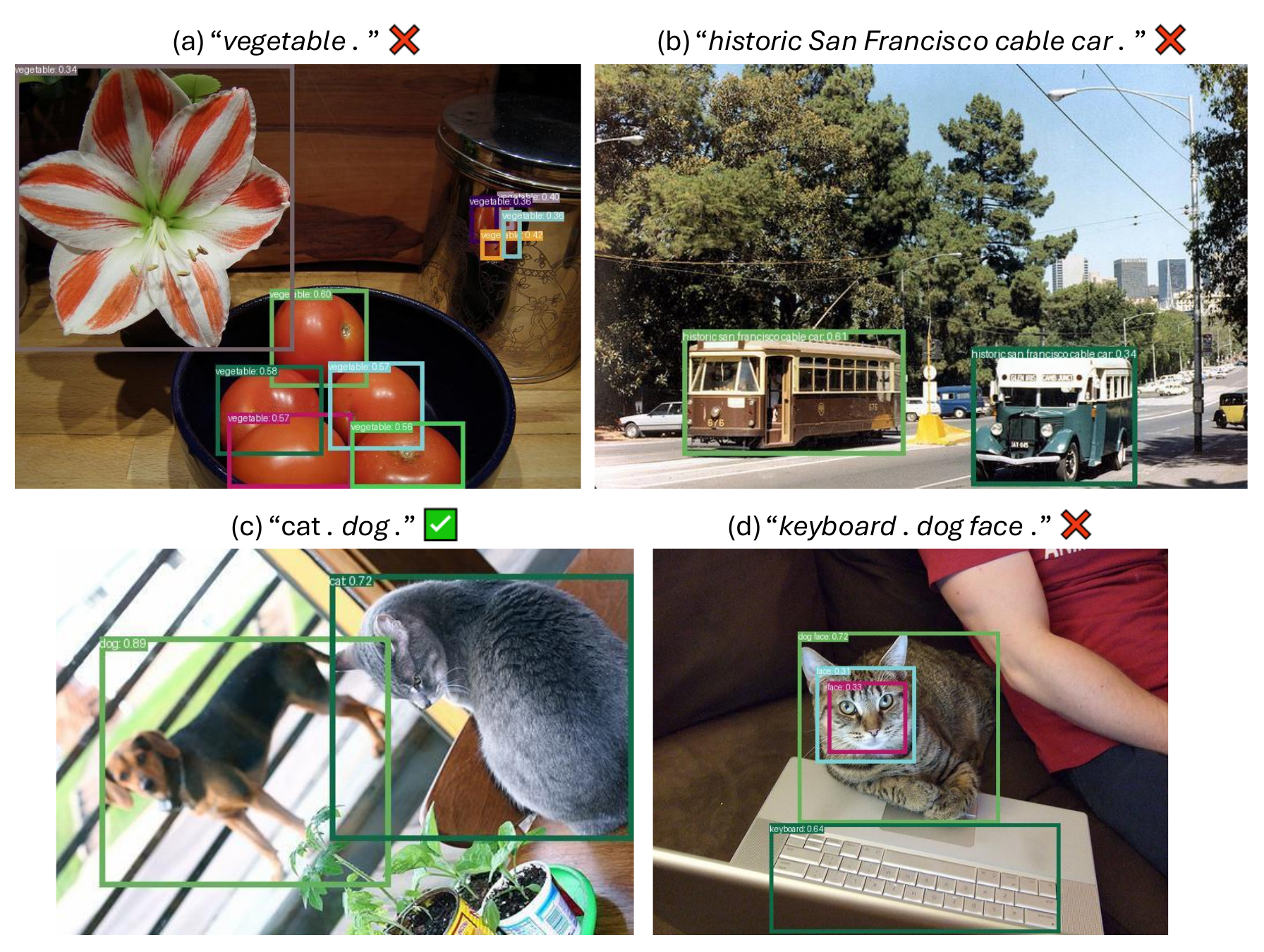}
  \caption{Failure cases. The model (a) is distracted by reflective regions of tomatoes; (b) fails to distinguish the target cable car from visually similar instances under fine-grained descriptions; (c) successfully differentiates between “cat” and “dog” under simple category prompts, but (d) hallucinates a “dog face” on a cat under compositional prompts.}
  \label{fig:fail}
\end{figure}

Although our \modelname{} improves semantic query generation and cross-modal alignment, we observe several failure cases. As shown in Fig.~\ref{fig:fail}, the model may be distracted by complex visual patterns such as reflections or cluttered backgrounds, leading to incorrect grounding results. Fine-grained open-vocabulary descriptions are also difficult when multiple visually similar objects coexist in the scene, as demonstrated by the cable car example. While the model can correctly distinguish coarse categories such as “cat” and “dog”, it may still hallucinate semantically related concepts under more compositional prompts (e.g., “dog face”). These examples highlight remaining challenges in language understanding.

\subsection{Future Work}

In the future, our work can be further extended toward improving the robustness of generative query modeling under more complex visual and textual inputs. Challenging visual conditions, such as occlusion, reflections, and cluttered backgrounds, may still interfere with accurate grounding, while fine-grained semantic distinctions between visually similar concepts remain difficult in open-vocabulary settings. We believe that the proposed generative query formulation provides a promising direction for reframing these challenges through continuous semantic modeling, especially when combined with stronger language reasoning, richer semantic supervision, and more structured generative objectives. Another promising direction is extending continuous query generation to broader multimodal perception tasks, such as video understanding and grounding.

\end{document}